\documentclass[journal]{IEEEtran}
\usepackage{subcaption}
\usepackage{amsmath,amsfonts}
\usepackage{algorithmic}
\usepackage{array}
\usepackage{cite}
\usepackage{textcomp}
\usepackage{stfloats}
\usepackage{url}
\usepackage{verbatim}
\usepackage{graphicx}
\usepackage{xcolor}
\usepackage{hyperref}
\usepackage{soul}
\usepackage{verbatim}
\usepackage{multirow}
\usepackage{booktabs}

\usepackage{orcidlink}
\usepackage{pifont}

\ifCLASSINFOpdf

\else

\fi

\begin{document}

\title{Co-AttenDWG: Co-Attentive Dimension-Wise Gating and Expert Fusion for Multi-Modal Offensive Content Detection}

\author{Md. Mithun Hossain \orcidlink{0009-0001-4883-1802}, Md.~Shakil~Hossain \orcidlink{0009-0009-1584-3282}, Sudipto~Chaki \orcidlink{0000-0002-7286-6722},  M. F. Mridha \orcidlink{0000-0001-5738-1631}, ~\IEEEmembership{Senior Member,~IEEE}

\thanks{Md. Mithun Hossain, Md. Shakil Hossain, and Sudipto Chaki are with the Department of Computer Science and Engineering, Bangladesh University of Business and Technology, Dhaka 1216, Bangladesh; e-mail: (mhosen751@gmail.com, shakilhosen3.1416@gmail.com, sudiptochakibd@gmail.com).}

\thanks{M. F. Mridha is with the Department of Computer Science, American International University-Bangladesh, Dhaka 1229, Bangladesh (email: firoz.mridha@aiub.edu).}

\thanks{Corresponding Author:  M. F. Mridha (e-mail: firoz.mridha@aiub.edu)}

\thanks{Manuscript received Month xx, 20xx; revised Month xx, 20xx.}}

\markboth{Journal of \LaTeX\ Class Files,~Vol.~14, No.~8, August~2015}%
{M. Mithun Hossain \MakeLowercase{\textit{et al.}}: Co-Attentive Dimension-Wise Gating and Expert Fusion}

\maketitle

\begin{abstract}
Multi-modal learning has emerged as a crucial research direction, as integrating textual and visual information can substantially enhance performance in tasks such as classification, retrieval, and scene understanding. Despite advances with large pre-trained models, existing approaches often suffer from insufficient cross-modal interactions and rigid fusion strategies, failing to fully harness the complementary strengths of different modalities. To address these limitations, we propose \textbf{Co-AttenDWG}, co-attention with dimension-wise gating, and expert fusion. Our approach first projects textual and visual features into a shared embedding space, where a dedicated co-attention mechanism enables simultaneous, fine-grained interactions between modalities. This is further strengthened by a dimension-wise gating network, which adaptively modulates feature contributions at the channel level to emphasize salient information. In parallel, dual-path encoders independently refine modality-specific representations, while an additional cross-attention layer aligns the modalities further. The resulting features are aggregated via an expert fusion module that integrates learned gating and self-attention, yielding a robust unified representation. Experimental results on the MIMIC and SemEval Memotion 1.0 datasets show that Co-AttenDWG achieves state-of-the-art performance and superior cross-modal alignment, highlighting its effectiveness for diverse multi-modal applications.
\end{abstract}

\textbf{\textit{Impact Statement--} The Co-AttenDWG architecture redefines multi-modal learning by overcoming limitations inherent in static fusion techniques. Integrating dual-path encoding, co-attention with dimension-wise gating, and advanced expert fusion, it dynamically harnesses complementary textual and visual cues in a unified embedding space. This approach significantly enhances cross-modal alignment and performance, as evidenced by state-of-the-art results on the MIMIC and SemEval Memotion datasets. By adaptively modulating feature contributions and refining representations, Co-AttenDWG not only boosts detection accuracy but also opens new avenues for intelligent, context-aware systems across domains such as content analysis, sentiment evaluation, and complex scene understanding. This breakthrough paves the way forward.}


%
\IEEEpeerreviewmaketitle

\section{Introduction}
\label{sec1}

Multi-modal learning has emerged as a transformative paradigm in artificial intelligence, driven by the necessity to integrate diverse data sources such as text, images, audio, and video to provide a holistic understanding of complex real-world scenarios~\cite{baltruvsaitis2018multimodal,ngiam2011multimodal,singh2025emogif}. This integration is particularly vital in tasks such as classification, sentiment analysis, and information retrieval, where the combination of complementary modalities reveals patterns and insights that remain hidden when each modality is processed independently~\cite{xu2015show}. Traditional methods typically process each modality separately and merge the results through simple concatenation or fixed-weight averaging~\cite{ngiam2011multimodal}. However, such basic fusion approaches often fail to capture the intricate interdependencies and correlations among modalities, resulting in suboptimal representations and limiting overall model performance~\cite{singh2024multiseao}.

\begin{figure}[ht]
    \centering
    \begin{subfigure}[b]{0.49\linewidth}
        \centering
        \includegraphics[width=\linewidth]{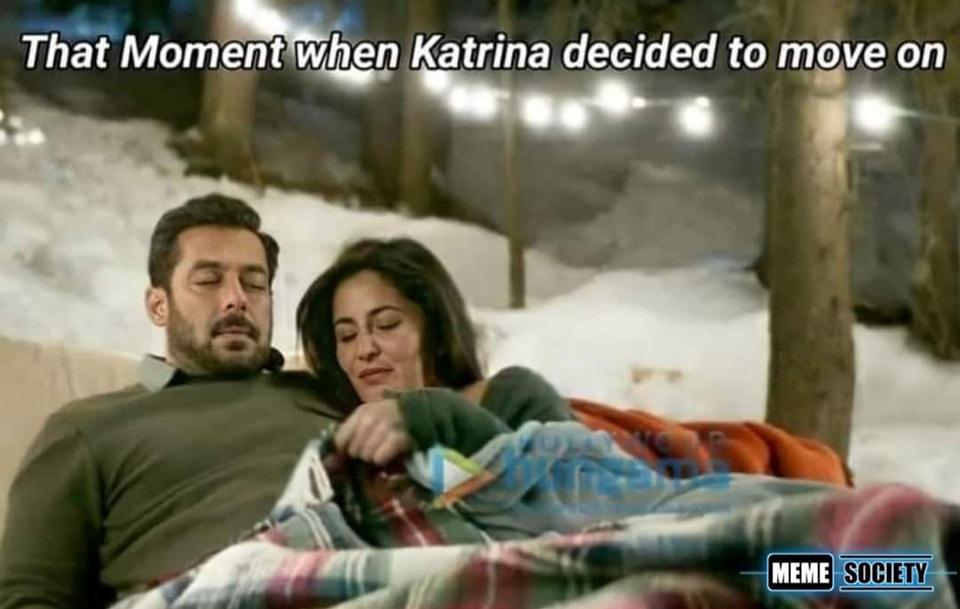} 
        \caption{\small Meme illustrating textual and visual interplay.}
        \label{subfig:meme_a}
    \end{subfigure}
    \hfill
    \begin{subfigure}[b]{0.49\linewidth}
        \centering
        \includegraphics[width=\linewidth]{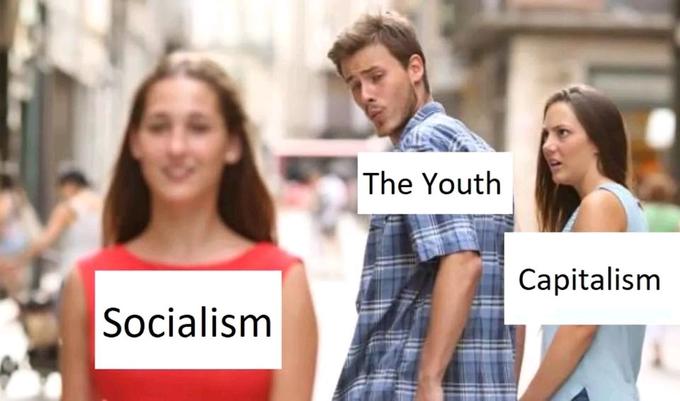} 
        \caption{\small Another meme combining images and text.}
        \label{subfig:meme_b}
    \end{subfigure}
    \caption{\small Examples of memes that combine textual cues with visual context, 
    illustrating the challenges of multi-modal integration. 
    Both examples demand nuanced interpretation of text, facial expressions, and background details.}
    \label{fig:meme_examples}
\end{figure}

Figure~\ref{fig:meme_examples} exemplifies the multifaceted challenges inherent in multi-modal data integration. In Figure~\ref{subfig:meme_a}, the meme combines textual humor with a richly nuanced visual context, where accurate interpretation depends not only on the literal meaning of the text but also on subtle visual cues such as facial expressions, gestures, and background elements that add layers of meaning and sentiment. Similarly, Figure~\ref{subfig:meme_b} presents a meme where layered textual cues interact with complex visual themes, including political symbolism and social context, requiring a sophisticated, fine-grained understanding of both modalities to fully grasp the intended message. These examples underscore the fundamental limitations of traditional static and simplistic fusion approaches that typically aggregate modalities without modeling their dynamic, context-dependent relationships. Such methods often fail to capture cross-modal dependencies and complementary information, resulting in suboptimal and sometimes misleading representations.

To address these challenges, recent research has leveraged powerful pre-trained models like BERT~\cite{devlin2019bert} and its multilingual and domain-adapted variants~\cite{liu2020multilingual, ruder2019unsupervised} for deep language understanding, alongside convolutional neural networks~\cite{krizhevsky2012imagenet,he2016deep} and vision transformers~\cite{dosovitskiy2020image} for visual feature extraction. While these architectures excel at generating robust, modality-specific embeddings, integrating them effectively remains a significant hurdle. Existing fusion strategies often rely on fixed or shallow combination mechanisms such as concatenation~\cite{ngiam2011multimodal}, early or late fusion~\cite{baltruvsaitis2018multimodal}, or simple attention mechanisms~\cite{lu2019vilbert}. However, these approaches inadequately align and reconcile the heterogeneous representations from different modalities. Recent advances propose more sophisticated cross-modal attention and co-attention mechanisms that dynamically model inter-modal interactions at multiple granularities~\cite{tan2019lxmert,li2019visualbert,chen2019uniter}, alongside gating networks that adaptively weigh features to suppress noise and highlight complementary signals~\cite{hossain2025dimension}. Transformer-based fusion modules~\cite{tsai2019multimodal} and graph neural networks for multimodal reasoning~\cite{li2019relation} have also shown promise in enhancing cross-modal alignment. These works highlight the growing consensus on the need for adaptive, context-aware fusion mechanisms capable of dynamically regulating cross-modal interactions at a fine-grained level. Such approaches selectively emphasize the most informative features from each modality depending on context, thereby improving interpretability and accuracy, particularly in complex tasks such as offensive content detection~\cite{hebert2024multi}, sentiment analysis~\cite{huang2023multimodal}, and multi-modal reasoning~\cite{kiela2019supervised}.

To address these shortcomings, we propose Co-AttenDWG, a novel multi-modal architecture that combines dual-path encoding with a co-attention mechanism enhanced by dimension-wise gating and advanced expert fusion. Our approach projects text and image features into a shared embedding space, where simultaneous, fine-grained cross-modal interactions occur via the co-attention mechanism. The dimension-wise gating network dynamically modulates channel-level feature contributions, selectively emphasizing the most informative components during fusion. We validate our approach on challenging datasets including MIMIC and SemEval Memotion 1.0, which require robust multi-modal comprehension. Experimental results demonstrate significant improvements in cross-modal alignment and state-of-the-art performance, illustrating the effectiveness and generalizability of our model.

The key contributions of this work are as follows:
\begin{itemize}
    \item We design a dual-path Co-AttenDWG architecture that robustly aligns and refines multi-modal representations.
    \item We introduce a dimension-wise gated co-attention mechanism to enable adaptive, fine-grained cross-modal interactions.
    \item We develop an expert fusion module that combines learned gating with self-attention to produce a unified, discriminative embedding.
\end{itemize}

The rest of this paper is organized as follows. \textcolor{blue}{Section \ref{literature}} iscusses related work in multi-modal offensive content detection and cross-modal fusion techniques. \textcolor{blue}{Section \ref{proposed_methodology}} outlines the \textbf{Co-AttenDWG} framework, including its key components and architectural design. \textcolor{blue}{Section \ref{RESULT}} presents our experiment, results, and provides a detailed analysis. \textcolor{blue}{Section \ref{sec5}} discusses the limitations of our study and prospective improvements that can be addressed in the future. Finally, \textcolor{blue}{Section \ref{CONCLUSIONS}} concludes the paper and highlights future research directions.

\section{Literature Review} \label{literature}
Recent advances in multi‐modal offensive content detection have increasingly focused on
uniting textual and visual cues to improve performance beyond traditional unimodal systems.
Early studies demonstrated that integrating features from pre‐trained language models and
computer vision architectures can significantly enhance detection accuracy. For example, Rana and Jha \cite{rana2022emotion} introduced a multimodal framework that fused BERT/ALBERT‐based text analysis with acoustic emotion cues in short videos, resulting in a notable reduction of false positives, particularly in discerning sarcasm from genuine hate speech. Likewise, Birhane et al. \cite{birhane2021multimodal} critically assessed large‐scale multimodal dataset revealing challenges related to explicit bias and noise while Suryawanshi et al. \cite{suryawanshi2020multimodal} showed that early fusion of text and image features in meme analysis yields improved detection results. In contrast, unimodal approaches \cite{paul2025multi, mu2024multimodal, huang2019image} that process either text or image data in isolation have consistently underperformed compared to models leveraging cross‐modal interactions, underscoring the necessity for more integrated methods.

Efficiently merging heterogeneous signals from different modalities is key to unlocking the full potential of multi‐modal systems \cite{ataei2022pars}. A variety of fusion strategies have been explored in the literature. Discriminative joint multi‐task frameworks, such as the one proposed by Zheng et al. \cite{zheng2024djmf}, utilize both intra‐ and inter‐task dynamics to enhance sentiment prediction. Chen et al. \cite{chen2023joint} further demonstrated that jointly fusing textual and visual features significantly improves classification accuracy by exploiting the complementary information inherent to each modality. While early and late fusion techniques \cite{abdullakutty2024decoding, zhu2023multimodal} offer a straightforward means for feature integration, they often suffer from modality‐specific information loss or misalignment. Hybrid approaches, which blend the strengths of both strategies \cite{gandhi2023multimodal, adel2023mmemor, zhou2023syntax, khan2023offensive} have provided more robust alternatives. Moreover, the introduction of cross‐attention mechanisms has allowed for fine‐grained interactions between visual and textual embeddings, as evidenced by recent studies from Mao et al. \cite{mao2025research} and Li et al. \cite{li2024multi}. Graph‐based fusion approaches \cite{hebert2024multi}, \cite{liang2024fusion} have also emerged, enabling models to capture contextual relationships through structured representations and addressing some limitations of simple concatenation schemes.
\begin{figure*}[h]
  \centering
  \includegraphics[width=\linewidth]{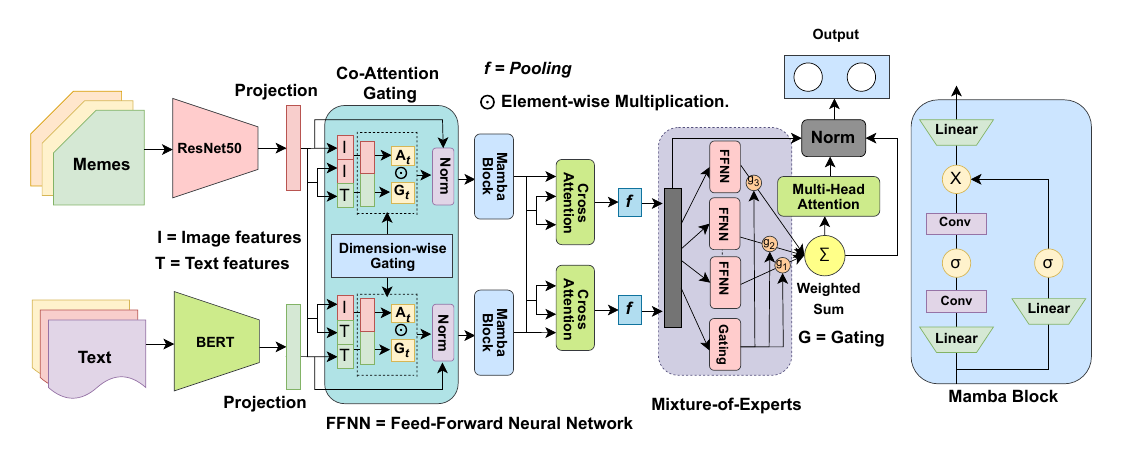}
  \caption{\small A high-level overview of our Co-AttenDWG architecture for multi-modal offensive content detection. The image branch (top) processes memes through a pre-trained CNN (ResNet50), extracting high-level visual features \(I\). Meanwhile, the text branch (bottom) encodes input sentences using a language model such as BERT, yielding textual features \(T\). Both sets of features are projected into a shared embedding space and enter the co-attention gating block, where dimension-wise gating adaptively emphasizes salient channels from each modality. The attention outputs, highlighted as \(A_i\) for images and \(A_t\) for text, pass into a mixture-of-experts fusion mechanism that combines relevant cross-modal cues. Next, the fused representation flows through the Mamba block, which integrates local convolutional operations and multi-head self-attention to refine context. Finally, the aggregated features are projected via linear layers to produce the final output, representing the predicted offensive content class. This design promotes dynamic cross-modal interactions and expert gating, enabling effective offensive content detection in both text and images.}
  \label{proposed_model}
\end{figure*}

Although pre‑trained models have advanced feature extraction from both text and image domains, current multi‑modal offensive content detection systems still rely on static fusion technique such as simple concatenation that inadequately capture the dynamic, context-dependent interplay between modalities. Existing attention‑based methods improve cross-modal alignment, yet they often neglect adaptive channel‑wise gating and expert-based fusion, limiting interpretability and robustness, particularly under noisy or ambiguous conditions. Models that rely solely on such static fusion techniques often fail to capture dynamic, structured cross-modal interactions in a bidirectional manner \cite{zheng2024djmf}. Furthermore, recent studies indicate that incorporating adaptive gating and expert fusion mechanisms substantially enhances a model’s ability to integrate complementary cues, resulting in improved performance and explainability \cite{chen2023joint}. To address these gaps, our proposed Co-AttenDWG model (see \textcolor{blue}{Figure \ref{proposed_model}}) projects text and image features into a shared space and employs bidirectional co‑attention coupled with a dimension‑wise gating mechanism to emphasize salient cues. An advanced expert fusion module  then adaptively combines modality-specific representations to enhance interpretability and performance, as further validated in our case studies. This dynamic, context-aware framework effectively overcomes the limitations of current static fusion strategies.

\section{Proposed Methodology}\label{proposed_methodology}

\textcolor{blue}{Figure \ref{proposed_model}} presents our proposed architecture, \textbf{Co-AttenDWG}, which addresses the challenges of multi-modal offensive content detection. Offensive content detection in multi-modal settings is challenging because text \(X_{\text{text}} \in \mathbb{R}^{B \times L}\) and images \(X_{\text{img}} \in \mathbb{R}^{B \times H \times W \times C}\) are processed through distinct pipelines that produce heterogeneous feature representations. For example, text is encoded using a model such as BERT \cite{devlin2019bert} that generates hidden states 
\(
H \in \mathbb{R}^{B \times L \times D_{\text{text}}}
\)
and extracts the representative [CLS] token 
\(
h_{\text{CLS}} = H[:,0,:] \in \mathbb{R}^{B \times D_{\text{text}}},
\)
which is then projected into a common embedding space to obtain 
\(
T \in \mathbb{R}^{B \times D}.
\)
Similarly, image features are extracted from a CNN such as ResNet50 \cite{he2016deep} to produce a feature vector 
\(
f \in \mathbb{R}^{B \times D_{\text{img}}},
\)
which is also projected into the same space as 
\(
I \in \mathbb{R}^{B \times D}.
\)
Traditional fusion strategies, such as simple concatenation \(F = [T; I]\), do not capture the dynamic, context-dependent cross-modal interactions needed for robust detection. To overcome these limitations, our method defines an adaptive fusion function \(\mathcal{F}(T, I)\) that produces a unified feature representation 
\(
E \in \mathbb{R}^{B \times D}
\)
by aligning heterogeneous modalities. This figure illustrates how our approach, through bidirectional fusion with co-attentive dimension-wise gating and expert fusion, emphasizes the most relevant features from each modality to enhance detection performance.

\subsection{Multi-Modal Feature Extraction and Projection}
\label{sec:feature-extraction}

In the text branch, we tokenize the input and process it with a pre-trained Transformer such as BERT \cite{devlin2019bert}, which has been demonstrated to excel at capturing long-range dependencies and contextual semantics in language. We extract the [CLS] hidden state as a global summary token, then project it into a common \(D\)-dimensional embedding space via a learned linear layer and reshape it into a token sequence for downstream fusion.
In the image branch, we employ a convolutional neural network (CNN) such as ResNet50 \cite{he2016deep} to extract hierarchical visual features. CNNs continue to be the de facto standard for picture encoding because of their effective inductive biases, such as weight sharing and local receptive fields, which allow for consistent training on sparse data and quick inference.  Similarly, these visual elements are molded into a pseudo-sequence and projected onto the common \(D\)-dimensional space.

\subsection{Bidirectional Fusion with Co-Attentive Dimension-Wise Gating}
We design a bidirectional fusion module to integrate features from text and image modalities while capturing fine-grained cross-modal interactions. Our approach first applies cross-modal attention \cite{vaswani2017attention} and then refines the outputs using a dimension-wise gating mechanism \cite{hu2018squeeze}.

\medskip
\noindent\textbf{Cross-Modal Co-Attention:} 
We let the text modality attend to the image modality by computing multi-head attention. Specifically, we use the text feature sequence \(T_{\text{seq}} \in \mathbb{R}^{B \times 1 \times D}\) as the query and the image feature sequence \(I_{\text{seq}} \in \mathbb{R}^{B \times 1 \times D}\) as both key and value. This yields:
\begin{equation}
  A_{t \rightarrow i} = \text{MHA}(Q = T_{\text{seq}},\, K = I_{\text{seq}},\, V = I_{\text{seq}}) \in \mathbb{R}^{B \times 1 \times D},
  \label{eq:att_text_to_img}
\end{equation}
which captures the image-informed features for the text modality. Similarly, we allow the image modality to attend to the text modality by using \(I_{\text{seq}}\) as the query and \(T_{\text{seq}}\) as both key and value:
\begin{equation}
  A_{i \rightarrow t} = \text{MHA}(Q = I_{\text{seq}},\, K = T_{\text{seq}},\, V = T_{\text{seq}}) \in \mathbb{R}^{B \times 1 \times D}.
  \label{eq:att_img_to_text}
\end{equation}

\medskip
\noindent\textbf{Dimension-Wise Gating:} 
After obtaining the attention outputs, we refine them using a channel-wise gating mechanism. For the text branch, we compute a gating weight:
\begin{equation}
  G_{t} = \sigma\left(W_{g,t}\, A_{t \rightarrow i} + b_{g,t}\right) \in \mathbb{R}^{B \times 1 \times D},
  \label{eq:gate_text}
\end{equation}
where \(\sigma\) is the sigmoid activation. This weight is then applied element-wise to the text attention output to obtain the gated text feature:
\begin{equation}
  \tilde{T} = G_{t} \odot A_{t \rightarrow i},
  \label{eq:gated_text}
\end{equation}
as shown in Equation~\eqref{eq:gated_text}. Similarly, for the image branch, we compute:
\begin{equation}
  G_{i} = \sigma\left(W_{g,i}\, A_{i \rightarrow t} + b_{g,i}\right) \in \mathbb{R}^{B \times 1 \times D},
  \label{eq:gate_image}
\end{equation}
and derive the gated image feature:
\begin{equation}
  \tilde{I} = G_{i} \odot A_{i \rightarrow t}.
  \label{eq:gated_image}
\end{equation}

\medskip
These steps align and enhance the features by emphasizing the most relevant information in each channel. The bidirectional attention, as defined in Equations~\eqref{eq:att_text_to_img} and \eqref{eq:att_img_to_text}, allows the modalities to inform each other, while the dimension-wise gating (Equations~\eqref{eq:gate_text}--\eqref{eq:gated_image}) selectively filters the features. This process improves the robustness of the subsequent fusion mechanism, enabling the model to dynamically capture cross-modal interactions and adaptively weight the contributions of each modality for more effective offensive content detection.

\begin{table}[ht]
\centering
\caption{\small Class Distributions Before and After Addressing Class Imbalance for MIMIC and Memotion Datasets}
\label{tab:class_distributions}
\begin{tabular}{lccc}
\toprule
\textbf{Dataset} & \textbf{Class Description}        & \textbf{Original Count} & \textbf{Balanced Count} \\
\midrule
\multirow{8}{*}{MIMIC} 
  & Non-Misogynistic   & 2497 & 2497 \\
  & Misogynistic       & 2409 & 2497 \\
  & Non-Humiliation    & 4537 & 4537 \\
  & Humiliation        & 369  & 4537 \\
  & Non-Objectification & 3462 & 3462 \\
  & Objectification    & 1444 & 3462 \\
  & Non-Prejudice      & 4032 & 4032 \\
  & Prejudice          & 874  & 4032 \\
\midrule
\multirow{4}{*}{Memotion} 
  & not\_offensive     & 2657 & 2657 \\
  & slight             & 2536 & 2657 \\
  & very\_offensive    & 1424 & 2657 \\
  & hateful\_offensive & 213  & 2657 \\
\bottomrule
\end{tabular}
\caption*{\footnotesize Note: The Memotion (Offensive Content) dataset mapping is \{"not\_offensive": 0, "slight": 1, "very\_offensive": 2, "hateful\_offensive": 3\}.}
\end{table}

\begin{table*}[ht]
\centering
\caption{\small Selected hyperparameters and data preparation details for Co-AttenDWG experiments.}
\label{tab:exp-settings}
\renewcommand{\arraystretch}{1.12}
\begin{tabular}{lcl}
\toprule
\textbf{Hyperparameter / Setting} & \textbf{Option} & \textbf{Best} \\
\midrule
{Datasets}                 & MIMIC / Memotion / Both & Both \\
{Language Coverage}        & English / Multilingual Hindi-English  & Both \\
{Multilingual Models}      & mBERT, XLM-RoBERTa & mBERT, XLM-RoBERTa \\
Optimizer                              & AdamW / SGD / RMSProp    & AdamW \\
Learning Rate                          & $1\text{e}{-5} / 2\text{e}{-5} / 5\text{e}{-5}$ & $2 \times 10^{-5}$ \\
Epochs                                 & 10 / 15 / 20             & 16 (Early Stop) \\
Learning Rate Scheduler                & None / Step / Dynamic    & Dynamic \\
Early Stopping                         & 3 / 5 / 7                & 3 \\
Attention Heads (Fusion)               & 4 / 8 / 12               & 8 \\
Self-Attention Heads (Refinement)      & 2 / 4 / 8                & 4 \\
MambaFormer Kernel Size                & 3 / 5 / 7                & 3 \\
MambaFormer Depth                      & 2 / 4 / 6                & 2 \\
Dropout Rate                           & 0.0 / 0.1 / 0.2          & 0.1 \\
Text Tokenizer                         & BERT / XLM-R / mBERT     & BERT / XLM-R \\
Image Normalization                    & Standard / MinMax / None & Standard (mean/std) \\
Image Size (MIMIC)                     & $160 \times 160$ / $200 \times 200$  & $200 \times 200$ px \\
Image Size (Memotion)                  & $128 \times 128$ / $160 \times 160$  & $160 \times 160$ px \\
\bottomrule
\end{tabular}
\end{table*}

\subsection{Dual-Path Encoding and Cross-Attention}
After refining the features with bidirectional co-attention and dimension-wise gating, we further enhance the representations through dual-path encoding and additional cross-attention mechanisms to refine and align the modalities before fusion.

\medskip
\noindent\textbf{Dual-Path Encoding:} 
The gated features are processed via MambaFormer-based encoder modules that combine self-attention and convolutional operations to capture both local and global context \cite{vaswani2017attention, hu2018squeeze}. For the text modality, we feed the gated image feature \(\tilde{I} \in \mathbb{R}^{B \times 1 \times D}\) into the text-to-image MambaFormer encoder to obtain a refined representation:
\begin{equation}
  Z_{t \rightarrow i} = \text{MambaFormerEncoder}(\tilde{I}) \in \mathbb{R}^{B \times 1 \times D},
  \label{eq:mb_encoder_text}
\end{equation}
which is then fused with the original text projection \(T_{\text{seq}}\) via element-wise addition:
\begin{equation}
  Z_{\text{text}} = Z_{t \rightarrow i} + T_{\text{seq}},
  \label{eq:fuse_text}
\end{equation}
as shown in Equation~\eqref{eq:fuse_text}. Similarly, for the image modality, we input the gated text feature \(\tilde{T} \in \mathbb{R}^{B \times 1 \times D}\) into the image-to-text MambaFormer encoder:
\begin{equation}
  Z_{i \rightarrow t} = \text{MambaFormerEncoder}(\tilde{T}) \in \mathbb{R}^{B \times 1 \times D},
  \label{eq:mb_encoder_img}
\end{equation}
and fuse it with the original image projection \(I_{\text{seq}}\) by element-wise addition:
\begin{equation}
  Z_{\text{img}} = Z_{i \rightarrow t} + I_{\text{seq}}.
  \label{eq:fuse_img}
\end{equation}

\medskip
\noindent\textbf{Cross-Attention:}
To further improve cross-modal alignment, an additional layer of cross-attention \cite{vaswani2017attention} is introduced. First, we let the text query \(T_{\text{seq}}\) attend to the image features \(I_{\text{seq}}\), computing:
\begin{equation}
  T_{\text{cross}} = \text{CrossAttn}(Q = T_{\text{seq}},\, K = I_{\text{seq}},\, V = I_{\text{seq}}) \in \mathbb{R}^{B \times 1 \times D},
  \label{eq:crossattn_text}
\end{equation}
which highlights image elements relevant to the text modality. Similarly, for the image modality, we compute:
\begin{equation}
  I_{\text{cross}} = \text{CrossAttn}(Q = I_{\text{seq}},\, K = T_{\text{seq}},\, V = T_{\text{seq}}) \in \mathbb{R}^{B \times 1 \times D},
  \label{eq:crossattn_img}
\end{equation}
capturing text elements informative for the image modality. The modality-specific representations are then updated by integrating these cross-attention outputs:
\begin{equation}
  Z_{\text{text}}^{\text{final}} = Z_{\text{text}} + T_{\text{cross}},
  \label{eq:final_text}
\end{equation}
\begin{equation}
  Z_{\text{img}}^{\text{final}} = Z_{\text{img}} + I_{\text{cross}},
  \label{eq:final_img}
\end{equation}
as detailed in Equations~\eqref{eq:final_text} and \eqref{eq:final_img}. 

\medskip
By employing dual-path encoding, we refine modality-specific features using the contextual modeling capacity of MambaFormer-based encoders. The additional cross-attention layers (Equations~\eqref{eq:crossattn_text} and \eqref{eq:crossattn_img}) further align the representations by integrating complementary information from each modality. This processing chain improves the overall quality of the extracted features, ensuring effective capture of complementary cues for subsequent fusion.

\subsection{Expert Fusion}

After aligning and refining the modality-specific features through dual-path encoding and additional cross-attention, we fuse the resulting experts into a single unified representation using an advanced expert fusion module.

\medskip
\noindent\textbf{Concatenation:}  
First, the final text and image representations are concatenated along the feature dimension:
\begin{equation}
  C = \big[Z_{\text{text}}^{\text{final}} ; Z_{\text{img}}^{\text{final}}\big] \in \mathbb{R}^{B \times 2D},
  \label{eq:concat}
\end{equation}
which combines complementary information from both modalities into a joint representation \(C\).

\medskip
\noindent\textbf{Fusion Network:}  
The concatenated features are then transformed using a feed-forward network with a non-linear activation \cite{vaswani2017attention}. The network produces an intermediate fused feature:
\begin{equation}
  F = \phi\left(W_f\, C + b_f\right) \in \mathbb{R}^{B \times D},
  \label{eq:fusion_net}
\end{equation}
where \(\phi(\cdot)\) (e.g., ReLU) introduces non-linearity. This step synthesizes the information from both text and image modalities.

\medskip
\noindent\textbf{Gating Weight Computation:}  
Next, we adaptively balance the modality contributions by computing gating weights. The gating network applies a linear transformation followed by a softmax function to \(C\) \cite{shazeer2017outrageously}:
\begin{equation}
  g = \text{softmax}\left(W_g\, C + b_g\right) \in \mathbb{R}^{B \times 2},
  \label{eq:gating_weights}
\end{equation}
where \(g = \left[g_{\text{text}},\, g_{\text{img}}\right]\) with each component representing the weight for the corresponding modality. The weighted expert sum is then computed as:
\begin{equation}
  S = g_{\text{text}} \odot Z_{\text{text}}^{\text{final}} + g_{\text{img}} \odot Z_{\text{img}}^{\text{final}} \in \mathbb{R}^{B \times D},
  \label{eq:weighted_sum}
\end{equation}
where \(\odot\) denotes element-wise multiplication.

\medskip
\noindent\textbf{Self-Attention Refinement:}  
To further refine the fused representation, we apply an additional self-attention layer \cite{vaswani2017attention}. First, we reshape \(S\) into a sequence of length one:
\begin{equation}
  S_{\text{seq}} \in \mathbb{R}^{1 \times B \times D},
  \label{eq:s_seq}
\end{equation}
and then compute:
\begin{equation}
  A = \text{MHA}(S_{\text{seq}}, S_{\text{seq}}, S_{\text{seq}}) \in \mathbb{R}^{1 \times B \times D},
  \label{eq:self_attn}
\end{equation}
after which \(A\) is reshaped back to \(\mathbb{R}^{B \times D}\).

\medskip
\noindent\textbf{Final Fusion:}  
The final unified multi-modal representation is obtained by summing the outputs of the fusion network, the weighted expert sum, and the self-attention refinement, followed by layer normalization \cite{ba2016layernorm}:
\begin{equation}
  E = \text{LayerNorm}\Big(F + S + A\Big) \in \mathbb{R}^{B \times D}.
  \label{eq:final_fusion}
\end{equation}
This final representation \(E\) encapsulates the complementary and dynamic interactions between the text and image features, preparing it for the classification stage.

\medskip
\noindent
Overall, the advanced expert fusion module leverages concatenation, adaptive gating via mixture-of-experts techniques \cite{shazeer2017outrageously}, and self-attention to integrate and refine modality-specific features. Equations~\eqref{eq:concat} through \eqref{eq:final_fusion} illustrate the step-by-step process that ensures the final representation robustly captures the essential information for effective offensive content detection.

\begin{table*}[ht]
\centering
\caption{\small Performance comparison of baselines and multimodal models on the MMIC and Memotion datasets. Best results in each column are \textbf{bolded}; second-best are \underline{underlined}. Inference time is measured on a single NVIDIA RTX 2060 12GB GPU, batch size 1.}
\label{tab:main-results}
\renewcommand{\arraystretch}{1.15}
\setlength{\tabcolsep}{4pt}
\begin{tabular}{lcccccccccccc}
\toprule
\multirow{2}{*}{\textbf{Model}} & 
\multirow{2}{*}{\textbf{Time (ms)}} &
\multicolumn{2}{c}{\textbf{Misogyny (\%)}} &
\multicolumn{2}{c}{\textbf{Objectification (\%)}} &
\multicolumn{2}{c}{\textbf{Prejudice (\%)}} &
\multicolumn{2}{c}{\textbf{Humiliation (\%)}} &
\multicolumn{2}{c}{\textbf{Memotion (\%)}} \\ 
\cmidrule(lr){3-4} \cmidrule(lr){5-6}  \cmidrule(lr){7-8} \cmidrule(lr){9-10} \cmidrule(lr){11-12}
 &  & \textbf{Acc} & \textbf{F1} & \textbf{Acc} & \textbf{F1} & \textbf{Acc} & \textbf{F1} & \textbf{Acc} & \textbf{F1} & \textbf{Acc} & \textbf{F1} \\
\midrule
mBERT~\cite{devlin2019bert}                  & 10.2 & 77.98 & 77.59 & 86.86 & 86.86 & 90.33 & 90.31 & 94.71 & 94.71 & --    & --    \\
BERT~\cite{devlin2019bert}                   & 10.2 & --   & --   & --   & --   & --   & --   & --   & --   & 78.60 & 78.82 \\
DistilBERT~\cite{sanh2019distilbert}         & 6.1  & --   & --   & --   & --   & --   & --   & --   & --   & 75.71 & 75.78 \\
VGG16~\cite{simonyan2014very}                & 12.5 & 84.58 & 84.58 & 92.51 & 92.51 & 95.04 & 95.04 & 97.23 & 97.23 & 76.83 & 76.78 \\
ResNet50~\cite{he2016deep}                   & 7.8  & 84.68 & 84.56 & 91.01 & 91.00 & 94.79 & 94.79 & 96.87 & 96.87 & 77.36 & 77.25 \\
EffNetV2~\cite{tan2021efficientnetv2}        & 6.5  & 81.68 & 81.65 & 90.90 & 90.90 & 94.23 & 94.23 & 96.99 & 96.99 & 62.14 & 61.20 \\
XLM-R~\cite{conneau2019unsupervised}         & 13.2 & 49.55 & 43.13 & 84.12 & 84.10 & 86.48 & 86.48 & 92.33 & 92.32 & --    & --    \\
\midrule
mBERT-VGG16                                 & 22.4 & 85.19 & 85.16 & 94.08 & 94.07 & 95.35 & 95.35 & 97.32 & 97.32 & --    & --    \\
mBERT-ResNet50                              & 17.5 & 85.99 & 85.98 & 94.51 & 94.51 & 96.09 & 96.09 & 97.28 & 97.28 & --    & --    \\
mBERT-Efficient                             & 16.3 & 83.08 & 82.90 & 93.65 & 93.65 & 95.29 & 95.29 & 97.56 & 97.56 & --    & --    \\
RoBERTa+ResNet50                            & 19.6 & {86.29} & {86.29} & 93.33 & 93.33 & 94.11 & 94.11 & 98.11 & 98.11 & --    & --    \\
RoBERTa+VGG16                               & 23.5 & 85.09 & 85.09 & 92.44 & 94.44 & 94.23 & 94.23 & 97.88 & 97.88 & --    & --    \\
RoBERTa+EffNetV2                            & 18.4 & 82.28 & 82.24 & 91.41 & 91.41 & 94.17 & 94.17 & 97.39 & 97.39 & --    & --    \\
mCLIP~\cite{radford2021learning}            & 16.0 & 85.79 & 85.78 & \underline{94.73} & \underline{94.73} & 95.06 & 95.05 & \underline{98.90} & \underline{98.90} & \underline{82.60} & \underline{82.66} \\
VisualBERT~\cite{li2019visualbert}          & 17.2 & \underline{86.39} & \underline{86.39} & {94.51} & {94.51} & \underline{97.02} & \underline{97.02} & \textbf{98.91} & \textbf{98.91} & {81.28} & {81.26} \\
ALBEF~\cite{li2021align}                    & 22.5 & 85.60 & 85.60 & 94.05 & 94.05 & 96.41 & 96.40 & 98.60 & 98.60 & 82.23 & 82.11 \\
BLIP~\cite{li2022blip}                      & 22.8 & 85.75 & 85.75 & 94.27 & 94.27 & 96.67 & 96.66 & 98.65 & 98.65 & 82.38 & 82.34 \\
BERT-ResNet50                               & 17.5 & --   & --   & --   & --   & --   & --   & --   & --   & 82.08 & 82.00 \\
BERT-Efficient                              & 16.3 & --   & --   & --   & --   & --   & --   & --   & --   & 79.21 & 78.94 \\
BERT-VGG16                                  & 22.4 & --   & --   & --   & --   & --   & --   & --   & --   & 81.10 & 81.00 \\
DistilBERT-ResNet50                         & 13.9 & --   & --   & --   & --   & --   & --   & --   & --   & 81.28 & 81.03 \\
DistilBERT-VGG16                            & 18.6 & --   & --   & --   & --   & --   & --   & --   & --   & 81.14 & 81.07 \\
DistilBERT-Efficient                        & 13.2 & --   & --   & --   & --   & --   & --   & --   & --   & 51.98 & 49.73 \\ \midrule
\textbf{Co-AttenDWG}                        & 31.1 & \textbf{87.19} & \textbf{87.16} & \textbf{94.80} & \textbf{94.80} & \textbf{97.15} & \textbf{97.15} & 98.80 & 98.80 & \textbf{84.29} & \textbf{84.26} \\
\textbf{Improvements} &  & \textcolor{blue}{+0.80$\uparrow$} & \textcolor{blue}{+0.77$\uparrow$}
                     & \textcolor{blue}{+0.07$\uparrow$} & \textcolor{blue}{+0.07$\uparrow$}
                     & \textcolor{blue}{+0.13$\uparrow$} & \textcolor{blue}{+0.13$\uparrow$}
                     & \textcolor{red}{-0.11$\downarrow$} & \textcolor{red}{-0.11$\downarrow$}
                     & \textcolor{blue}{+1.69$\uparrow$} & \textcolor{blue}{+1.60$\uparrow$} \\

\bottomrule
\end{tabular}
\caption*{\footnotesize Note: RoBERTa = XLM-RoBERTa \cite{conneau2019unsupervised}, Efficient = EfficientNetV2 \cite{tan2021efficientnetv2}. Inference time is measured on an NVIDIA RTX 2060 12GB GPU, batch size 1. Memotion (Offensive Content) \\}
\end{table*}

\begin{table*}[ht]
\centering
\caption{\small Combinatorial ablation study of the Co-AttenDWG model on the MMIC and Memotion datasets. Each row disables or modifies one or more major components. Best results are \textbf{bolded}.}
\label{tab:ablation}
\renewcommand{\arraystretch}{1.15}
\setlength{\tabcolsep}{3.5pt}
\begin{tabular}{lcccccccccc}
\toprule
\multirow{2}{*}{\textbf{Model Variant}} &
\multicolumn{2}{c}{\textbf{Misogyny (\%)}} &
\multicolumn{2}{c}{\textbf{Objectification (\%)}} &
\multicolumn{2}{c}{\textbf{Prejudice (\%)}} &
\multicolumn{2}{c}{\textbf{Humiliation (\%)}} &
\multicolumn{2}{c}{\textbf{Memotion (\%)}} \\ 
\cmidrule(lr){2-3} \cmidrule(lr){4-5} \cmidrule(lr){6-7} \cmidrule(lr){8-9} \cmidrule(lr){10-11}
& \textbf{Acc ($\uparrow$)} & \textbf{F1 ($\uparrow$)} & \textbf{Acc ($\uparrow$)} & \textbf{F1 ($\uparrow$)} & \textbf{Acc ($\uparrow$)} & \textbf{F1 ($\uparrow$)} & \textbf{Acc ($\uparrow$)} & \textbf{F1 ($\uparrow$)} & \textbf{Acc ($\uparrow$)} & \textbf{F1 ($\uparrow$)} \\
\midrule
w/o EF               & 83.58 & 83.56 & 93.86 & 93.86 & 88.03 & 87.98 & 97.80 & 97.80 & 80.90 & 80.84 \\
w/o CA               & 85.59 & 85.57 & 93.50 & 93.50 & 91.82 & 91.81 & 98.10 & 98.10 & 81.19 & 81.01 \\
w/o XA               & 83.58 & 83.57 & 94.01 & 94.01 & 92.75 & 92.74 & 97.60 & 97.60 & 79.12 & 79.26 \\
w/o MF               & 84.98 & 84.95 & 93.29 & 93.29 & 90.95 & 90.93 & 97.85 & 97.85 & 81.19 & 81.31 \\
w/o FF               & 85.27 & 85.25 & 93.78 & 93.78 & 91.40 & 91.38 & 97.90 & 97.90 & 80.81 & 80.69 \\
w/o EF+MF            & 83.10 & 83.06 & 93.05 & 93.04 & 88.89 & 88.86 & 97.40 & 97.40 & 79.55 & 79.62 \\
w/o CA+XA            & 82.21 & 82.20 & 93.13 & 93.12 & 88.41 & 88.39 & 97.10 & 97.10 & 78.66 & 78.59 \\
w/o EF+CA+MF         & 81.40 & 81.32 & 92.57 & 92.55 & 87.07 & 87.03 & 96.80 & 96.80 & 77.12 & 77.04 \\
2Heads               & 85.66 & 85.64 & 93.99 & 93.99 & 91.78 & 91.77 & 97.50 & 97.50 & 80.63 & 80.58 \\
w/o MF+FF            & 82.71 & 82.68 & 93.00 & 93.00 & 89.18 & 89.17 & 97.55 & 97.55 & 79.89 & 79.87 \\
\textbf{Full (Co-AttenDWG)} & \textbf{87.19} & \textbf{87.16} & \textbf{94.80} & \textbf{94.80} & \textbf{97.15} & \textbf{97.15} & \textbf{98.80} & \textbf{98.80} & \textbf{84.29} & \textbf{84.26} \\
\bottomrule
\end{tabular}
\caption*{\footnotesize EF = ExpertFusion; CA = Co-Attention; XA = Cross-Attention; MF = MambaFormer; FF = Fine-grained Fusion; 2Heads = Reduced Attention Heads (4$\rightarrow$2). Memotion (Offensive Content)}
\end{table*}

\begin{table*}[ht]
\centering
\caption{\small Impact of core architectural hyperparameters (number of experts, cross-attention heads, co-attention heads, MambaFormer kernel size, depth, dropout, pixel value, and learning rate) on macro F1 (\%) for each label. Results are on the MIMIC and Memotion validation sets. Best per column are in \textbf{bold}.}
\label{tab:arch-ablation-labels}
\renewcommand{\arraystretch}{1.15}
\setlength{\tabcolsep}{3pt}
\begin{tabular}{cccccccccccccc}
\toprule
\textbf{\# Experts} & \textbf{Cross-} & \textbf{Co-} & \textbf{Kernel} & \textbf{Depth} & \textbf{Dropout} & \textbf{Pixel} & \textbf{LR} & 
& \textbf{Misogyny (\%)} & \textbf{Object. (\%)} & \textbf{Prejudice (\%)} & \textbf{Humil. (\%)} & \textbf{Memotion (\%)} \\
 & \textbf{Attn} & \textbf{Attn} & \textbf{Size} & & & \textbf{Value} &  & & & & & & \\
\midrule
4 & 4 & 4 & 7  & 4 & 0.10  & 224 & $2\times10^{-5}$   & & 85.11 & 93.00 & 95.01 & 97.21 & 80.98 \\
8 & 4 & 4 & 9  & 4 & 0.15 & 200 & $3\times10^{-5}$   & & 85.69 & 93.22 & 95.33 & 97.48 & 81.31 \\
8 & 8 & 4 & 5  & 6 & 0.20  & 160 & $1\times10^{-5}$   & & 86.13 & 93.60 & 95.68 & 97.93 & 81.98 \\
8 & 8 & 8 & 7  & 8 & 0.05 & 128 & $5\times10^{-5}$   & & 86.79 & 94.10 & 96.92 & 98.51 & 84.01 \\
8 & 8 & 8 & 11 & 6 & 0.10  & 200 & $4\times10^{-5}$   & & 86.52 & 94.00 & 96.30 & 98.34 & 83.45 \\
8 & 8 & 8 & 5  & 3 & 0.30  & 128 & $1.5\times10^{-5}$ & & 86.35 & 94.22 & 96.20 & 98.22 & 83.01 \\
\textbf{8} & \textbf{8} & \textbf{4} & \textbf{3}  & \textbf{2} & \textbf{0.10}  & \textbf{200/160} & \textbf{$2\times10^{-5}$} &  & \textbf{87.16}  & \textbf{94.80}  & \textbf{97.15}  & \textbf{98.80}  & \textbf{84.26} \\
8 & 8 & 8 & 7  & 4 & 0.00  & 224 & $2.5\times10^{-5}$ &  & 86.88 & 94.35 & 96.75 & 98.08 & 83.99 \\
\bottomrule
\end{tabular}
\caption*{\footnotesize Object. = Objectification, Humil. = Humiliation, Memotion (Offensive Content)}
\end{table*}

\begin{table*}[ht]
\centering
\caption{\small Co-AttenDWG performance using different backbone models on the MIMIC and Memotion datasets (Accuracy and F1 in \%). Best results are \textbf{bolded}. Memotion (Offensive Content)}
\label{tab:backbone-ablation}
\renewcommand{\arraystretch}{1.15}
\setlength{\tabcolsep}{4pt}
\begin{tabular}{llcccccccccc}
\toprule
\multicolumn{2}{c}{\textbf{Backbone}} & 
\multicolumn{2}{c}{\textbf{Misogyny (\%)}} &
\multicolumn{2}{c}{\textbf{Objectification (\%)}} &
\multicolumn{2}{c}{\textbf{Prejudice (\%)}} &
\multicolumn{2}{c}{\textbf{Humiliation (\%)}} &
\multicolumn{2}{c}{\textbf{Memotion (\%)}} \\
\cmidrule(lr){1-2} \cmidrule(lr){3-4} \cmidrule(lr){5-6} \cmidrule(lr){7-8} \cmidrule(lr){9-10}\cmidrule(lr){11-12}
\textbf{Text} & \textbf{Image} &
\textbf{Acc} & \textbf{F1} &
\textbf{Acc} & \textbf{F1} &
\textbf{Acc} & \textbf{F1} &
\textbf{Acc} & \textbf{F1} &
\textbf{Acc} & \textbf{F1} \\
\midrule
mBERT & ResNet50          & 85.29 & 85.26 & 94.08 & 94.08 & 96.09 & 96.09 & 98.80 & 98.80 & --    & --    \\
mBERT & VGG16             & 85.29 & 86.83 & 94.73 & 94.73 & 95.23 & 95.23 & 98.80 & 98.80 & --    & --    \\
mBERT & EfficientNetV2    & 85.28 & 85.27 & 94.30 & 94.30 & 94.07 & 94.05 & 96.99 & 96.99 & --    & --    \\
\textbf{XLM-RoBERTa} & \textbf{ResNet50} & \textbf{87.79} & \textbf{87.83} & \textbf{94.80} & \textbf{94.80} & \textbf{97.15} & \textbf{97.15} & \textbf{98.80} & \textbf{98.80} & -- & -- \\
XLM-RoBERTa & VGG16         & 86.17 & 86.82 & 91.91 & 91.91 & 97.02 & 97.02 & 98.80 & 98.80 & --    & --    \\
XLM-RoBERTa & EfficientNetV2& 82.58 & 82.86 & 90.40 & 90.40 & 92.25 & 92.25 & 97.01 & 97.01 & --    & --    \\
\midrule
BERT & ResNet50           & --    & --    & --    & --    & --    & --    & --    & --    & \textbf{84.29} & \textbf{84.26} \\
BERT & VGG16              & --    & --    & --    & --    & --    & --    & --    & --    & 83.66 & 83.61 \\
BERT & EfficientNetV2     & --    & --    & --    & --    & --    & --    & --    & --    & 81.19 & 81.01 \\
DistilBERT & ResNet50      & --    & --    & --    & --    & --    & --    & --    & --    & 81.81 & 81.81 \\
DistilBERT & VGG16         & --    & --    & --    & --    & --    & --    & --    & --    & 82.09 & 82.07 \\
DistilBERT & EfficientNetV2& --    & --    & --    & --    & --    & --    & --    & --    & 79.97 & 80.01 \\
\bottomrule
\end{tabular}
\end{table*}

\subsection{Classification}

After obtaining the unified multi-modal representation \(E \in \mathbb{R}^{B \times D}\), a linear classifier is applied to map this representation to class logits for \(C\) classes. The classifier transforms the multi-modal features into logits, which are then normalized using the softmax function to yield predicted class probabilities. The final predicted class for each sample is determined by selecting the class with the highest probability. The entire model is trained end-to-end using cross-entropy loss, comparing the predicted probabilities with the true class labels. Optimization is performed using the AdamW optimizer with an appropriate learning rate schedule. During training, both the pre-trained encoders (for text and image) and the fusion and classification layers are fine-tuned to learn effective cross-modal interactions that facilitate robust offensive content detection.

\section{Experiment and Result Analysis}\label{RESULT}
\subsection{Datasets}
We evaluate our Co-AttenDWG model on two publicly available datasets that present diverse and challenging scenarios for multi-modal offensive content detection. The first dataset, SemEval-2020 Memotion Analysis 1.0 \cite{sharma2020task}, is widely used as a benchmark for offensive meme classification (Offensive) on social media platforms. It captures the complex range of harmful material in meme visual and word combinations with rich annotations across four levels of offensiveness: not offensive, slightly offensive, highly offensive, and hateful offensive. The second dataset, MIMIC: Misogyny Identification in Multimodal Internet Content \cite{singh2024mimic}, is specifically designed to address misogynistic behavior in Hindi-English code-mixed multimodal posts, a setting that introduces unique linguistic and cultural challenges for joint text-image understanding.
MIMIC comprises four distinct classification tasks targeting related yet conceptually different forms of offensive content: Misogynistic, Humiliation, Objectification, and Prejudice. Notably, the MIMIC dataset is multilingual, featuring code-mixed Hindi-English posts, which introduces additional challenges related to language diversity and code-switching in multi-modal contexts. Each of these categories requires the model to recognize subtle cues in both textual and visual modalities, making MIMIC a comprehensive testbed for evaluating fine-grained multi-label multi-modal classification in multilingual and code-mixed scenarios. The dataset is naturally imbalanced, with significant disparities in the distribution of positive and negative examples across each class. To mitigate this, we apply upsampling techniques to balance the classes, ensuring that the model receives sufficient training examples from minority classes, thereby improving generalization and robustness. \textcolor{blue}{Table~\ref{tab:class_distributions}} summarizes the original and balanced class counts for all these categories across both datasets.

Together, the Memotion and MIMIC datasets provide a rigorous evaluation framework for our architecture, enabling us to assess its capability to handle complex multi-modal inputs, code-mixed language, and subtle offensive content distinctions across both multilingual and English language content in culturally diverse contexts. This comprehensive evaluation demonstrates the effectiveness and adaptability of Co-AttenDWG in real-world multi-modal offensive content detection scenarios.

\subsection{Implementation Details}
We implement our Co-AttenDWG model using Python 3.12.1 and PyTorch 2.0.1 on an NVIDIA RTX 2060 GPU with 16~GB of RAM. Our optimization strategy employs the AdamW optimizer with a fixed learning rate of \(2 \times 10^{-5}\), training for 20 epochs while leveraging a dynamic learning rate scheduler that adjusts the rate during training for improved convergence. The model architecture is configured with 8 attention heads in the fusion modules and 4 heads in the self-attention refinement layer. The MambaFormer encoders are set with a kernel size of 3, a depth of 2 layers, and a dropout rate of 0.1 applied uniformly across all modules. All components, including the pre-trained text encoders (BERT and XLM-RoBERTa) and the image encoder (ResNet50), are fine-tuned end-to-end to maximize cross-modal feature alignment. The detailed hyperparameter settings and data preparation options are summarized in \textcolor{blue}{Table~\ref{tab:exp-settings}}.

In terms of data preparation, we carefully clean and normalize both text and images. Text is tokenized using the BERT tokenizer, while images are resized and normalized according to dataset-specific requirements. For the MIMIC dataset which contains multilingual Hindi-English code-mixed data images are resized to \(200 \times 200\) pixels to preserve detail, whereas the SemEval Memotion 1.0 dataset images are resized to \(160 \times 160\) pixels due to resource constraints. We partition both datasets into 80\% training and 20\% testing splits and evaluate model performance using test accuracy and macro F1-score as primary metrics. To address the inherent class imbalance particularly prominent in minority offensive categories we apply upsampling strategies to balance the class distributions, ensuring equitable representation during training. These carefully chosen hyperparameter settings and robust preprocessing techniques enable Co-AttenDWG to effectively capture fine-grained cross-modal interactions and demonstrate superior performance on complex multi-modal offensive content detection tasks.

\begin{figure}[ht]
\centering
\includegraphics[width=0.9\linewidth]{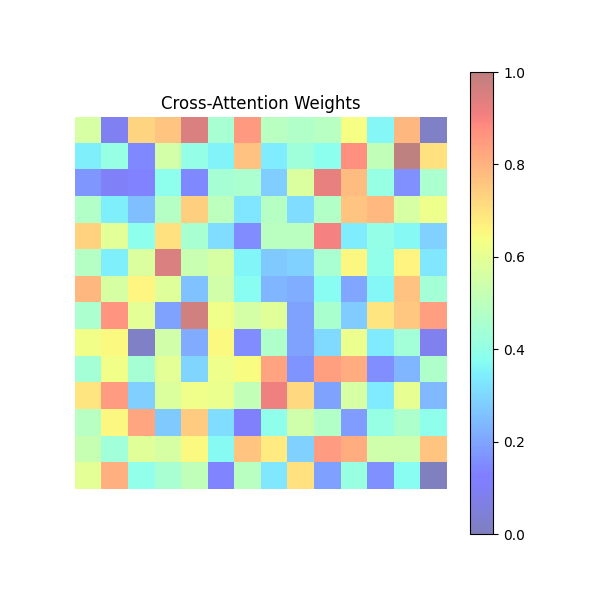}
\caption{\small Cross-attention weight distribution in our Co-AttenDWG architecture. Each cell represents the attention magnitude from a text token to a visual feature. Warmer colors indicate higher attention, and cooler colors indicate lower attention, with the scale ranging from 0 (lowest) to 1 (highest).}
\label{attention_weights}
\end{figure}

\begin{figure}[ht]
    \centering
    \begin{subfigure}[b]{0.9\linewidth}
        \centering
        \includegraphics[width=\linewidth]{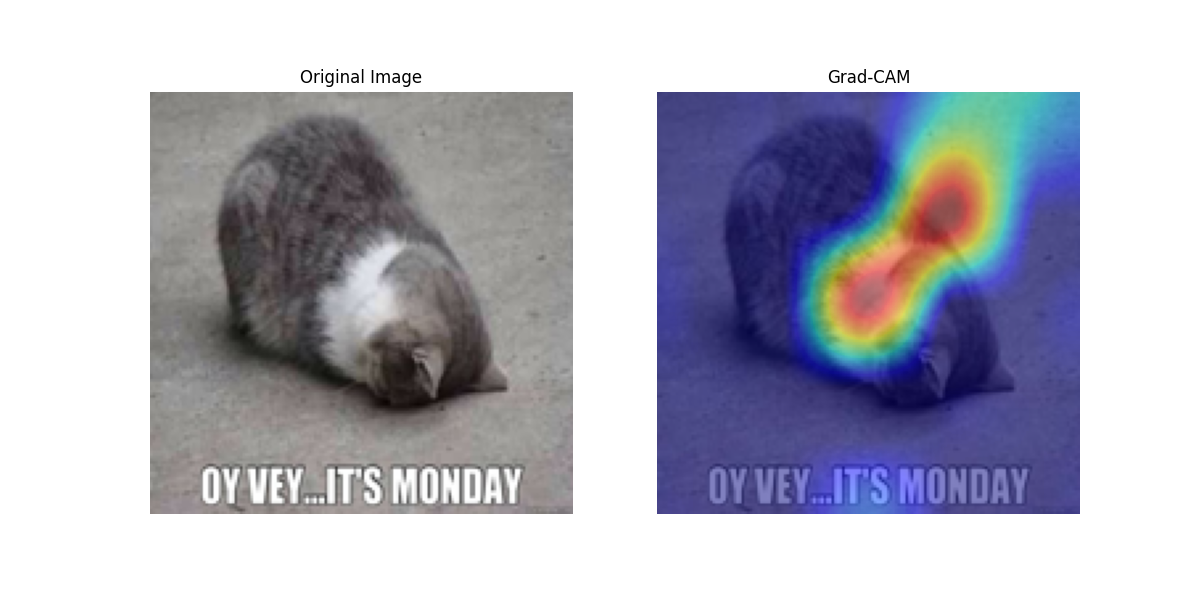}
        \caption{ Original cat meme (left) and Grad-CAM heatmap (right) highlighting salient regions for the classifier.}
        \label{gradcam_1}
    \end{subfigure}
    \hfill
    \begin{subfigure}[b]{0.9\linewidth}
        \centering
        \includegraphics[width=\linewidth]{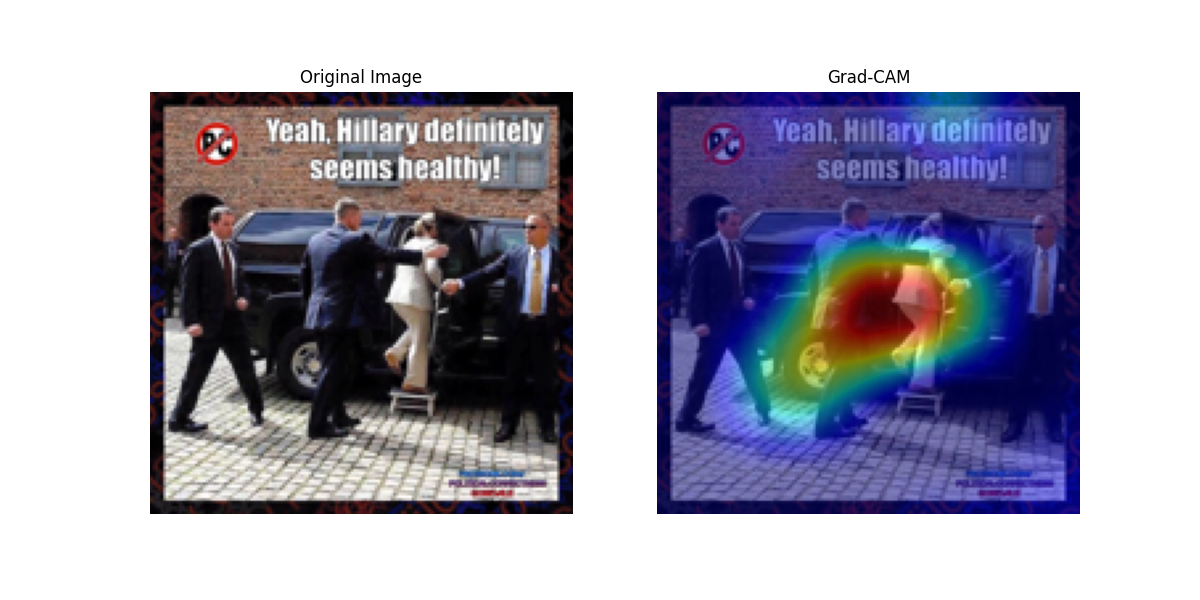}
        \caption{ Original political meme (left) and Grad-CAM heatmap (right). The model focuses on the main subject and text.}
        \label{gradcam_2}
    \end{subfigure}
    \caption{\small Examples of Grad-CAM visualizations demonstrating which regions of the images the model deems most salient. Figure (a) shows a “cat meme” context, while Figure (b) depicts a political scene. In both cases, the heatmap on the right reveals how the Co-AttenDWG classifier interprets key visual clues.
    }
    \label{gradcam}
\end{figure}

\begin{figure*}[ht]
    \centering
    \begin{subfigure}[b]{0.24\linewidth}
        \centering
        \includegraphics[width=\linewidth]{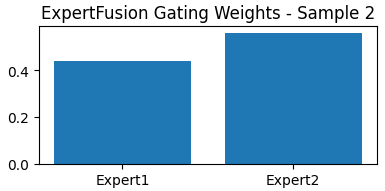}
        \caption{ExpertFusion Gating Weights - Sample 2}
        \label{1}
    \end{subfigure}
    \hfill
    \begin{subfigure}[b]{0.24\linewidth}
        \centering
        \includegraphics[width=\linewidth]{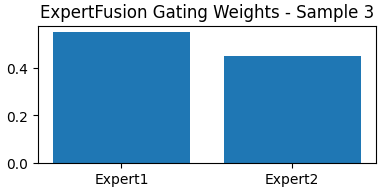}
        \caption{ExpertFusion Gating Weights - Sample 3}
        \label{2}
    \end{subfigure}
    \hfill
    \begin{subfigure}[b]{0.24\linewidth}
        \centering
        \includegraphics[width=\linewidth]{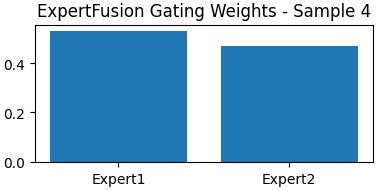}
        \caption{ExpertFusion Gating Weights - Sample 4}
        \label{3}
    \end{subfigure}
        \begin{subfigure}[b]{0.24\linewidth}
        \centering
        \includegraphics[width=\linewidth]{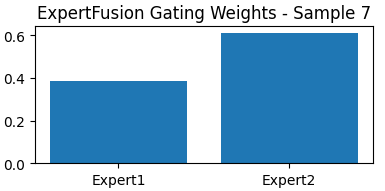}
        \caption{ExpertFusion Gating Weights - Sample 7}
        \label{4}
    \end{subfigure}
    
    \caption{\small Bar plots illustrating the gating weights assigned to each expert for different samples in the ExpertFusion module. 
    Each Figure (a, b, c, d) corresponds to a distinct sample, showcasing how the gating mechanism adapts to different inputs.}
    \label{fig:expertfusion_gating_all}
\end{figure*}

\begin{figure*}[ht]
    \centering
    \begin{subfigure}[b]{0.24\textwidth}
        \includegraphics[width=\textwidth]{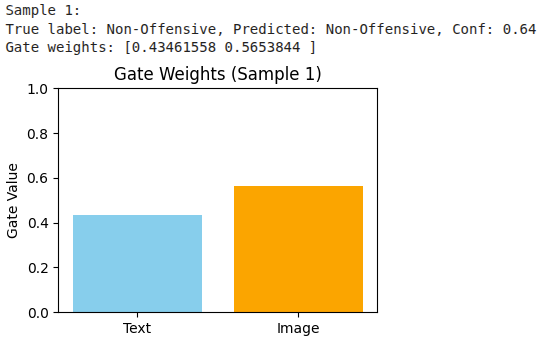}
        \includegraphics[width=\textwidth]{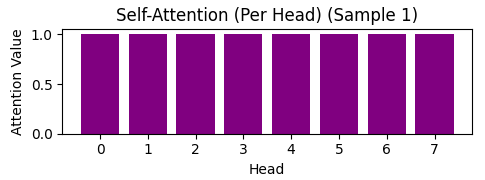}
        \caption{Sample 1: Balanced gate, slightly favors image; moderate confidence.}
        \label{fine_1}
    \end{subfigure}
    \begin{subfigure}[b]{0.24\textwidth}
        \includegraphics[width=\textwidth]{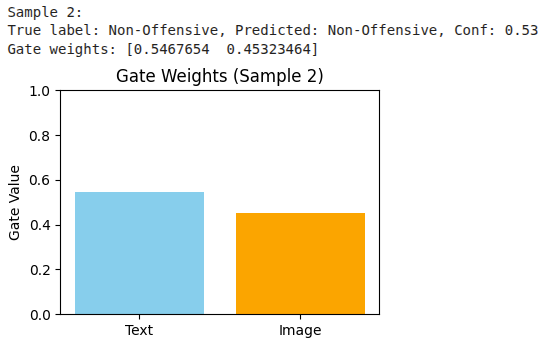}
        \includegraphics[width=\textwidth]{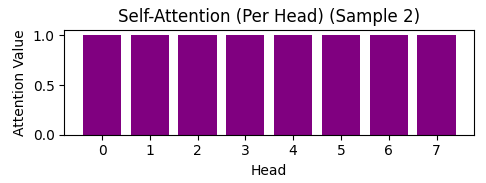}
        \caption{Sample 2: Low confidence, gate favors text.}
        \label{fine_2}
    \end{subfigure}
    \begin{subfigure}[b]{0.24\textwidth}
        \includegraphics[width=\textwidth]{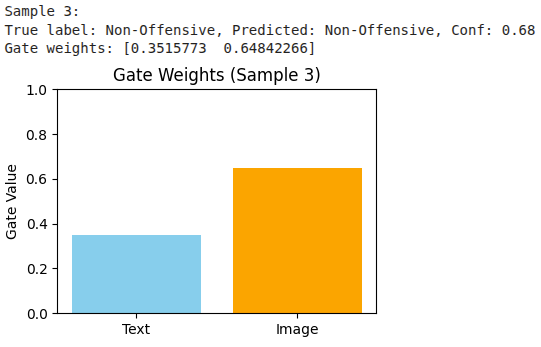}
        \includegraphics[width=\textwidth]{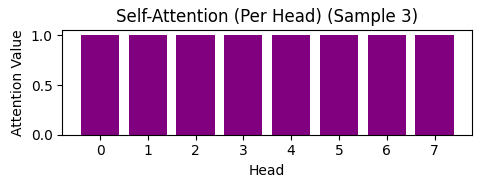}
        \caption{Sample 3: High confidence, strong image gating.}
        \label{fine_3}
    \end{subfigure}
    \begin{subfigure}[b]{0.24\textwidth}
        \includegraphics[width=\textwidth]{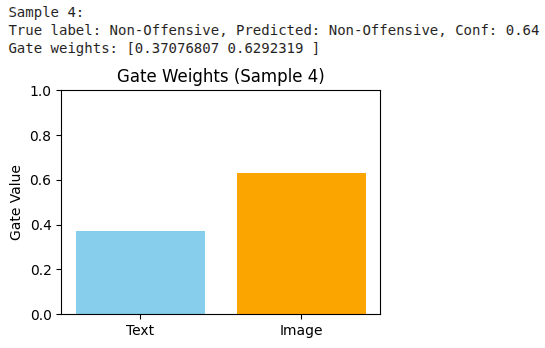}
        \includegraphics[width=\textwidth]{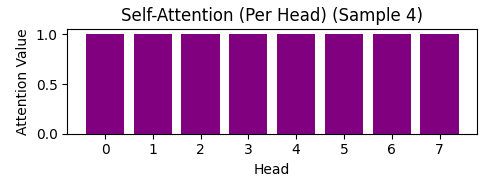}
        \caption{Sample 4: Confident prediction, attention sharp on image.}
        \label{fine_4}
    \end{subfigure}
    \caption{\small Self-attention heatmap visualizations for selected samples from the fine-grained interpretability analysis. Each subfigure illustrates how the model distributes attention between modalities and across sequence/image regions for the given example.}
    \label{fig:self-attn-finegrained}
\end{figure*}

\subsection{Baseline Comparison}

\textcolor{blue}{Table~\ref{tab:main-results}} presents an extensive performance comparison between our proposed Co-AttenDWG model and a wide range of baseline and state-of-the-art multimodal models evaluated on the MIMIC and SemEval Memotion datasets. The evaluation includes key offensive content detection categories such as Misogyny, Objectification, Prejudice, Humiliation, and Offensive Content from the Memotion dataset, reporting both accuracy and F1 scores to comprehensively capture performance. Co-AttenDWG consistently outperforms all baseline models across the majority of categories, establishing new state-of-the-art results with accuracy and F1 scores of 87.19\% and 87.16\% for Misogyny detection, 94.80\% for both metrics in Objectification, 97.15\% in Prejudice, and 84.29\% accuracy alongside 84.26\% F1 on the Memotion offensive content detection task. These results demonstrate Co-AttenDWG’s exceptional ability to capture subtle and complex cross-modal interactions between textual and visual modalities, crucial for nuanced offensive content understanding. In the Humiliation category, VisualBERT emerges as the highest-performing baseline, achieving 98.91\% in both accuracy and F1 scores. Co-AttenDWG closely follows with a very competitive 98.80\%, trailing by a marginal 0.11 percentage points. Notably, other strong vision-language models such as mCLIP, ALBEF, and BLIP also perform well in this category, with mCLIP and VisualBERT setting a high bar for multi-modal understanding. Despite the slight dip in the Humiliation metric, Co-AttenDWG surpasses these models substantially in all other categories, reflecting its balanced and robust performance profile.
The inference time of Co-AttenDWG is measured at 31.1 milliseconds per sample on an NVIDIA RTX 2060 12GB GPU with a batch size of one. This is competitive considering the advanced architectural components, such as dual-path encoders, co-attention with dimension-wise gating, and expert fusion modules, which collectively contribute to its superior accuracy and fine-grained modeling capacity.

Overall, these results underscore the strength of Co-AttenDWG in effectively integrating and refining multi-modal signals for offensive content detection, outperforming current vision-language models and baseline architectures. The model’s ability to generalize across diverse and culturally nuanced datasets like MIMIC and Memotion highlights its potential applicability in real-world scenarios requiring sensitive and precise offensive content moderation.

\subsection{Ablation Study}
\textcolor{blue}{Table~\ref{tab:ablation}} presents a comprehensive combinatorial ablation study evaluating the impact of removing or modifying key components of the Co-AttenDWG model on performance across the MMIC and Memotion datasets. Each variant disables or alters one or more major modules, such as ExpertFusion (EF), Co-Attention (CA), Cross-Attention (XA), MambaFormer (MF), and Fine-grained Fusion (FF), or reduces the number of attention heads. The results show consistent performance degradation across all tasks when any of these components are removed, confirming their individual contributions to the overall model effectiveness. Notably, the full Co-AttenDWG model achieves the highest accuracies and F1 scores, reaching up to 87.19\% accuracy on Misogyny and 98.80\% on Humiliation, demonstrating robust multi-modal learning capabilities. The largest performance drops occur when multiple critical modules are removed simultaneously, such as the combination of ExpertFusion, Co-Attention, and MambaFormer, which substantially lowers results across all evaluated categories. This ablation analysis validates the importance of each architectural element in capturing fine-grained, cross-modal interactions essential for state-of-the-art offensive content detection and sentiment understanding.

\subsection{Findings}

\textcolor{blue}{Table~\ref{tab:arch-ablation-labels}} presents an ablation study analyzing the impact of core architectural hyperparameters on the Co-AttenDWG model’s performance across multiple labels in the MIMIC and Memotion validation sets. The hyperparameters explored include the number of experts, cross-attention heads, co-attention heads, MambaFormer kernel size and depth, dropout rate, input image resolution (pixel value), and learning rate. The results demonstrate that increasing the number of experts and attention heads generally enhances performance, with the best configuration employing 8 experts, 8 cross-attention heads, and 4 co-attention heads. A smaller kernel size of 3 and a shallow MambaFormer depth of 2 layers, combined with a moderate dropout rate of 0.10, also contribute to optimal results. Image resolution plays a role, with the best setting utilizing a combination of 200 and 160 pixels depending on the dataset. Learning rate tuning around \(2 \times 10^{-5}\) further stabilizes training. This optimal setting yields the highest macro F1 scores across all labels: 87.16\% for Misogyny, 94.80\% for Objectification, 97.15\% for Prejudice, 98.80\% for Humiliation, and 84.26\% for the Memotion dataset. These findings underscore the importance of carefully balancing architectural complexity and regularization to maximize multi-label multimodal classification performance.

\textcolor{blue}{Table~\ref{tab:backbone-ablation}} presents an ablation study evaluating the impact of different backbone combinations on the performance of the Co-AttenDWG model across the MIMIC and Memotion datasets. We explore three text encoders (mBERT, XLM-RoBERTa, and BERT variants) paired with three image backbones (ResNet50, VGG16, and EfficientNetV2) to assess how backbone selection affects multi-modal offensive content detection. Across all MIMIC sub-tasks, including Misogyny, Objectification, Prejudice, and Humiliation, the results show that the multilingual XLM-RoBERTa text encoder together with the ResNet50 image backbone achieves the best overall performance, with accuracies and F1 scores consistently above 87\% and 94\%, respectively. Notably, this combination also maintains high effectiveness on the Memotion dataset. While mBERT backbones also deliver strong results, particularly with ResNet50 and VGG16, they generally perform slightly below the top-performing XLM-RoBERTa models. EfficientNetV2 backbones show comparatively lower results, especially in the Humiliation category. The BERT and DistilBERT variants, though effective on Memotion, lack reported results for several MIMIC sub-tasks, reflecting possible limitations in handling code-mixed multilingual data. These findings underline the critical role of backbone selection in multi-modal architectures and support the use of powerful, multilingual text encoders and strong visual backbones to maximize performance in complex offensive content detection tasks.

\begin{table}[ht]
\centering
\caption{\small True Negatives (TN), False Positives (FP), False Negatives (FN), and True Positives (TP) for all binary and multiclass tasks.}
\begin{tabular}{lcccc}
\toprule
\textbf{Task / Class}     & \textbf{TN} & \textbf{FP} & \textbf{FN} & \textbf{TP} \\
\midrule
Humiliation             & 923  & 10   & 0    & 882  \\
Misogyny                & 455  & 49   & 85   & 410  \\
Objectification         & 639  & 41   & 34   & 671  \\
Prejudice               & 763  & 64   & 16   & 770  \\ \midrule
Memotion (Offensive Content) & & & &  \\ \midrule
not\_offensive          & 1446 & 150  & 148  & 382  \\
slight                  & 1462 & 114  & 172  & 378  \\
very\_offensive         & 1492 & 121  & 80   & 433  \\
hateful\_offensive      & 1578 & 15   & 0    & 533  \\
\bottomrule
\end{tabular}
\label{Error_analysis}
\end{table}

\begin{figure*}[ht]
    \centering
    \begin{subfigure}[b]{0.24\linewidth}
        \centering
        \includegraphics[width=\linewidth]{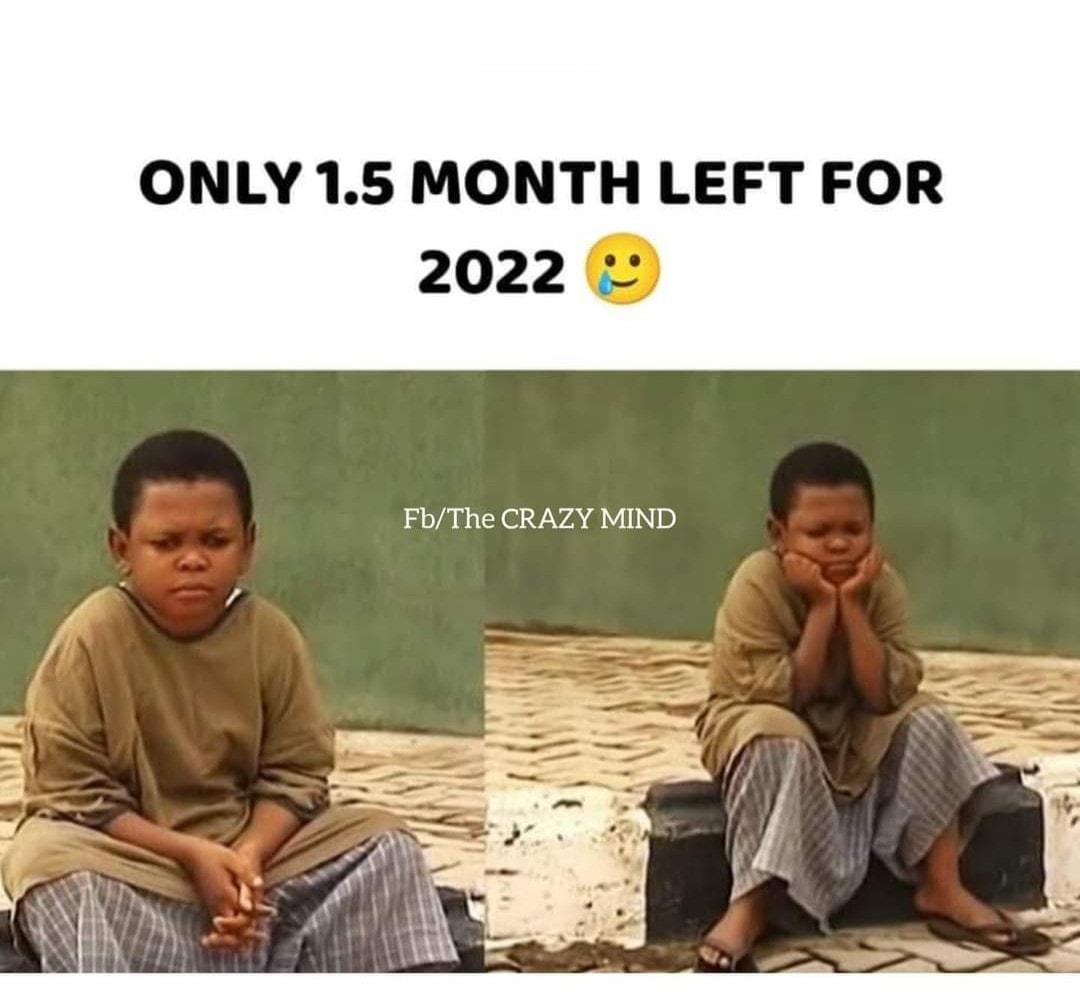}
        \caption{\small Memotion sample (True label: not\_offensive)\\
        mCLIP - (\ding{55})\\
        VisualBERT - (\ding{51})\\
        ALBEF - (\ding{51})\\
        BLIP - (\ding{51})\\
        Co-AttenDWG - (\ding{51})}
        \label{sub1}
    \end{subfigure}
    \hfill
    \begin{subfigure}[b]{0.24\linewidth}
        \centering
        \includegraphics[width=\linewidth]{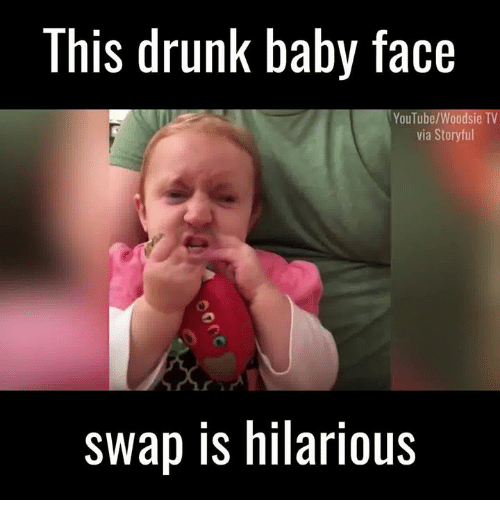}
        \caption{\small Memotion sample (True label: hateful\_offensive)\\
        mCLIP - (\ding{51})\\
        VisualBERT - (\ding{51})\\
        ALBEF - (\ding{51})\\
        BLIP - (\ding{51})\\
        Co-AttenDWG - (\ding{51})}
        \label{sub2}
    \end{subfigure}
    \hfill
\begin{subfigure}[b]{0.24\linewidth}
    \centering
    \includegraphics[width=\linewidth]{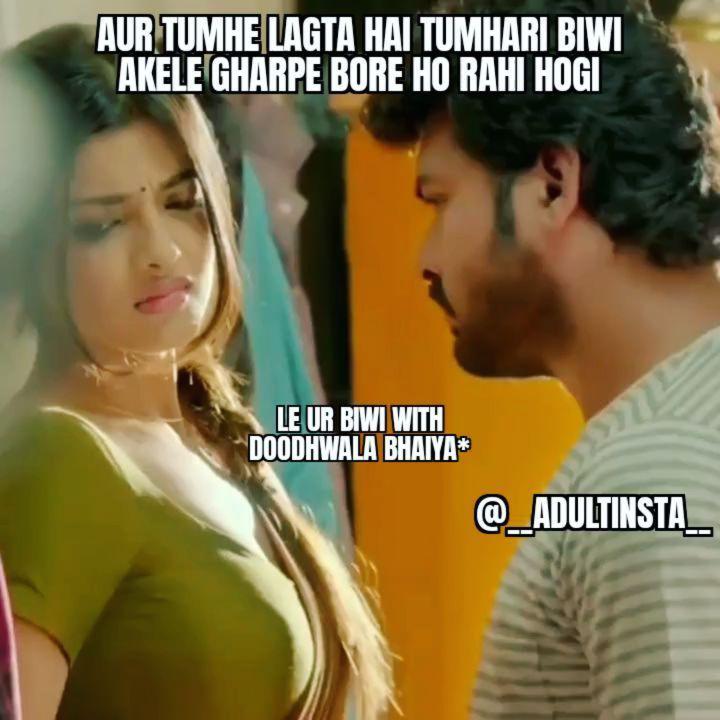}
    \caption{\small MIMIC sample with true labels: misogyny, prejudice, objectification, humiliation. \\
    mCLIP - misogyny (\ding{51}), prejudice (\ding{51}), objectification (\ding{55}), humiliation (\ding{51}).\\
    VisualBERT - misogyny (\ding{51}), prejudice (\ding{51}), objectification (\ding{51}), humiliation (\ding{51}).\\
    ALBEF - misogyny (\ding{51}), prejudice (\ding{55}), objectification (\ding{51}), humiliation (\ding{51}).\\
    BLIP - misogyny (\ding{51}), prejudice (\ding{55}), objectification (\ding{51}), humiliation (\ding{51}).\\
    Co-AttenDWG - misogyny (\ding{51}), prejudice (\ding{51}), objectification (\ding{51}), humiliation (\ding{51}).}
    \label{sub3}
\end{subfigure}
\hfill
\begin{subfigure}[b]{0.24\linewidth}
    \centering
    \includegraphics[width=\linewidth]{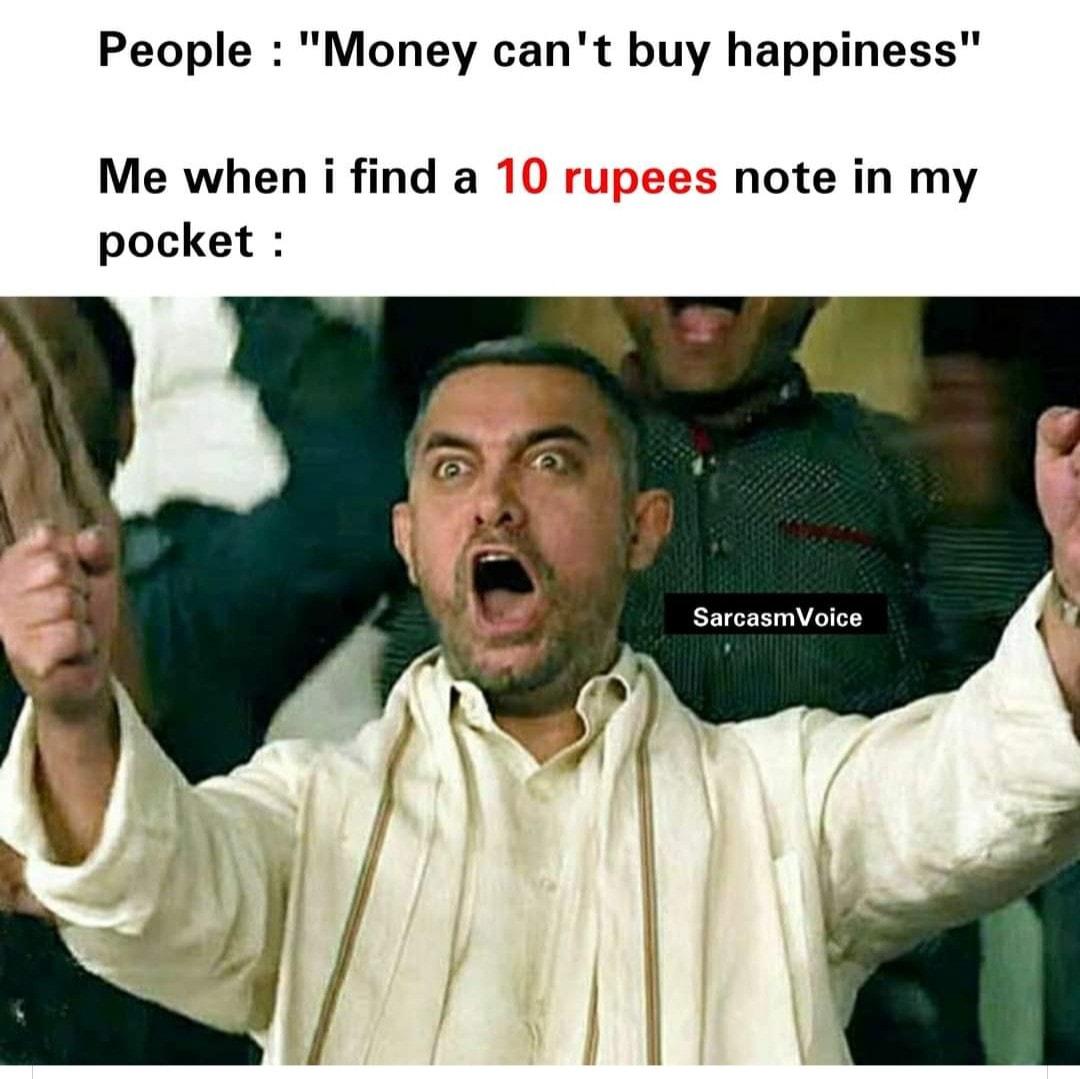}
    \caption{\small MIMIC sample with true labels: misogyny, prejudice, objectification, humiliation. \\
    mCLIP - misogyny (\ding{51}), prejudice (\ding{55}), objectification (\ding{55}), humiliation (\ding{51}).\\
    VisualBERT - misogyny (\ding{55}), prejudice (\ding{55}), objectification (\ding{51}), humiliation (\ding{51}).\\
    ALBEF - misogyny (\ding{55}), prejudice (\ding{51}), objectification (\ding{51}), humiliation (\ding{51}).\\
    BLIP - misogyny (\ding{51}), prejudice (\ding{51}), objectification (\ding{51}), humiliation (\ding{51}).\\
    Co-AttenDWG - misogyny (\ding{51}), prejudice (\ding{51}), objectification (\ding{51}), humiliation (\ding{51}).}
    \label{sub4}
\end{subfigure}
\caption{\small Case study examples on offensive content detection. The top two subfigures show Memotion samples with true labels "not\_offensive" (a) and "hateful\_offensive" (b). In (a), mCLIP fails to detect offensive content (\ding{55}), while VisualBERT, ALBEF, BLIP, and Co-AttenDWG correctly classify the sample (\ding{51}); in (b), all models correctly predict the offensive content (\ding{51}). The other two subfigures depict MIMIC samples with true labels including misogyny, prejudice, objectification, and humiliation. In (c), all models correctly classify all labels (\ding{51}), whereas in (d), VisualBERT incorrectly classifies misogyny and prejudice (\ding{55}) but correctly identifies objectification and humiliation (\ding{51}), while mCLIP, ALBEF, BLIP, and Co-AttenDWG correctly classify all labels (\ding{51}).}

    \label{case}
\end{figure*}

\subsection{Interpretability}
\textcolor{blue}{Figure~\ref{attention_weights}} presents a detailed heatmap visualization of the cross-attention weights learned by the Co-AttenDWG architecture. Each cell in the heatmap quantifies the attention strength from a specific textual token to a corresponding visual feature, with warmer colors signifying higher attention values and cooler colors indicating lower values on a normalized scale from 0 to 1. This visualization demonstrates how the model dynamically aligns relevant textual cues with semantically meaningful regions within the image, effectively capturing intricate cross-modal dependencies. Complementing this, \textcolor{blue}{Figure~\ref{gradcam}} shows Grad-CAM visualizations that highlight salient image regions influencing the classifier’s predictions. Specifically, subfigure (\ref{gradcam_1}) depicts a “cat meme” where the heatmap emphasizes key visual elements aligned with textual content, whereas subfigure (\ref{gradcam_2}) illustrates a political meme, indicating attention over both the principal subject and embedded textual information. Collectively, these visualizations validate the model’s capacity to interpret and fuse multi-modal features in a meaningful and interpretable manner. Furthermore, \textcolor{blue}{Figure~\ref{fig:expertfusion_gating_all}} presents bar plots that illustrate the gating weights assigned by the ExpertFusion module to different experts across diverse samples. The shift in gating distributions from a preponderance of one expert to more balanced weightings among several experts demonstrates the module's adaptability and flexibility in adjusting the impact of different feature extractors according to input data.  Such dynamic expert weighting is necessary to improve the model's overall representational capability and integrate complementary multi-modal information in an efficient manner.

\textcolor{blue}{Figure~\ref{fig:self-attn-finegrained}} presents an in-depth examination of the fine-grained interpretability of the Co-AttenDWG model through a series of self-attention heatmap visualizations for four representative test samples. \textcolor{blue}{Subfigure \ref{fine_1}} (Sample 1) shows the gating network allocating balanced attention weights to both text and image modalities, with a slight emphasis on visual features, reflecting moderate confidence and demonstrating the model’s capacity to integrate complementary signals harmoniously. \textcolor{blue}{Subfigure \ref{fine_2}} (Sample 2) illustrates a low-confidence instance where the gating mechanism predominantly favors the textual modality, indicating that the model appropriately relies more heavily on language features when visual cues are ambiguous or less informative. \textcolor{blue}{Subfigure \ref{fine_3}} (Sample 3) depicts a high-confidence prediction characterized by a strong gating bias toward the image modality, highlighting the model’s ability to prioritize salient visual information when it serves as a more definitive classification indicator. Lastly, \textcolor{blue}{Subfigure \ref{fine_4}} (Sample 4) demonstrates a confident prediction accompanied by sharply focused self-attention on specific spatial regions within the image, evidencing the model’s proficiency in localizing and attending to critical visual cues that substantively contribute to its decision-making process. Together, these subfigures reveal the dynamic, context-sensitive fusion strategy employed by Co-AttenDWG, illustrating how the interplay between gating and attention mechanisms adapts to the input data to improve interpretability and classification robustness in complex multi-modal scenarios.

\subsection{Error Analysis}
\textcolor{blue}{Table~\ref{Error_analysis}} presents detailed counts of true negatives (TN), false positives (FP), false negatives (FN), and true positives (TP) for both the MIMIC and Memotion datasets, providing an in-depth analysis of classification performance. For the MIMIC dataset, the Humiliation class exhibits excellent classification performance, with 923 true negatives and 882 true positives, and minimal errors, indicating the model’s strong ability to identify this category. In contrast, the Misogyny class shows higher confusion, with 49 false positives and 85 false negatives, highlighting challenges in correctly distinguishing misogynistic content due to subtle textual and visual cues. Objectification and Prejudice classes demonstrate moderate misclassification rates, with a balanced distribution of false positives and false negatives, suggesting overlapping features in multimodal inputs contribute to classification ambiguity. For the Memotion dataset, which focuses on offensive content intensity levels, the “not\_offensive” and “hateful\_offensive” categories show relatively strong separability with higher true negatives and true positives and fewer misclassifications. However, intermediate classes such as “slight” and “very\_offensive” have notable misclassification rates, reflecting the difficulty in distinguishing nuanced differences between similar offensive intensities. These observations suggest that while the model effectively handles broad category distinctions, fine-grained discrimination among closely related classes remains challenging, underscoring the potential for improved feature extraction and more consistent annotation to enhance multi-class classification performance.

\textcolor{blue}{Figure~\ref{case}} presents four illustrative examples demonstrating the efficacy of the Co-AttenDWG architecture for offensive content detection in multimodal memes, alongside comparisons with mCLIP and VisualBERT. In \textcolor{blue}{Figure (\ref{sub1})}, a Memotion sample labeled as "not\_offensive" is examined, where both VisualBERT and Co-AttenDWG correctly classify the sample, whereas mCLIP fails to detect its non-offensive nature. This indicates that Co-AttenDWG and VisualBERT possess greater sensitivity to subtle non-offensive cues in multimodal content. In contrast, \textcolor{blue}{Figure (\ref{sub2})} shows a "hateful\_offensive" Memotion meme that all three models classify correctly, reflecting their robustness when identifying clearly offensive content. Shifting focus to the MIMIC dataset, \textcolor{blue}{Figure (\ref{sub3})} depicts a misogynistic post that all models accurately recognize, signifying consistent detection capabilities for misogyny across architectures. However, in \textcolor{blue}{Figure (\ref{sub4})}, another misogynistic MIMIC example reveals divergent model behaviors: while mCLIP and Co-AttenDWG correctly identify the offensive content, VisualBERT misclassifies the instance, underscoring Co-AttenDWG’s improved generalization in challenging or ambiguous cases. Collectively, these examples elucidate how Co-AttenDWG effectively integrates textual and visual information to maintain high classification accuracy across diverse, real-world scenarios, surpassing or matching state-of-the-art alternatives in both non-offensive and offensive content detection.

\section{Limitations and Future Works}
\label{sec5}
While the Co-AttenDWG architecture achieves notable improvements in multimodal offensive content detection, several limitations remain. First, although the model effectively fuses textual and visual modalities, it can face challenges when processing highly ambiguous, context-dependent content, especially within code-mixed or low-resource language scenarios. The use of fixed pre-trained backbones such as BERT and ResNet may limit adaptability to emerging linguistic patterns and novel visual meme formats, potentially impacting robustness over time. Second, the current model primarily targets single-label or multi-class classification tasks and does not explicitly model hierarchical or multi-label dependencies where overlapping offensive content categories coexist. This restricts the model’s ability to fully capture the nuanced relationships among different forms of offense. Third, while upsampling addresses class imbalance, it may inadvertently cause overfitting or bias toward synthetic samples, underscoring the need for more sophisticated imbalance mitigation strategies, including focal loss or data augmentation approaches.

For future work, incorporating Vision Transformer (ViT) models could significantly enhance visual representation learning due to their superior ability to capture long-range dependencies and global contextual information in images. Extending the architecture to handle additional modalities such as audio and video would further broaden applicability in multimedia-rich social platforms. Furthermore, developing dynamic fusion techniques that adaptively modulate interactions based on input complexity, as well as incorporating continual learning paradigms, can improve model generalization over time and across domains. The model could also be adapted and evaluated on other multimodal tasks such as sentiment analysis, hate speech detection, or misinformation classification, to validate its generalizability and robustness across diverse applications. Finally, integrating explainability mechanisms will be crucial to increase transparency, foster trust, and support ethical deployment in real-world content moderation systems.

\section{Conclusions}\label{CONCLUSIONS}
In this work, we proposed Co-AttenDWG, a novel multimodal architecture that effectively integrates textual and visual information through dual-path encoding, co-attention with dimension-wise gating, and expert fusion. Our approach dynamically captures fine-grained cross-modal interactions, enabling robust alignment of heterogeneous features. Extensive experiments on the MIMIC and SemEval Memotion datasets demonstrated that Co-AttenDWG consistently outperforms state-of-the-art baselines, achieving superior accuracy and F1 scores across multiple offensive content detection categories. The qualitative analyses further reveal the model’s ability to focus on semantically meaningful regions in both modalities, highlighting its interpretability and adaptability to diverse, culturally nuanced contexts. While challenges remain in handling ambiguous content and class imbalance, our results establish a strong foundation for future multimodal research. The proposed framework can be extended to incorporate advanced visual encoders such as Vision Transformers and to support richer multimodal inputs beyond text and images. Overall, Co-AttenDWG advances the field of multimodal offensive content detection by providing a powerful, flexible, and interpretable solution capable of addressing complex real-world scenarios.

\bibliographystyle{IEEEtran}
\bibliography{sample}

\end{document}